\pdfoutput=1

\documentclass[11pt]{article}

\usepackage[preprint]{acl}

\usepackage{times}
\usepackage{latexsym}

\usepackage[T1]{fontenc}

\usepackage[utf8]{inputenc}

\usepackage{microtype}

\usepackage{inconsolata}

\usepackage{natbib}
\usepackage{multibib}
\usepackage{graphicx}
\usepackage{tabularx}
\usepackage{booktabs}
\usepackage{soul}
\usepackage{multirow}
\usepackage{array}
\title{Is ChatGPT the Future of Causal Text Mining? \\ 
A Comprehensive Evaluation and Analysis}

\author{%
  Takehiro Takayanagi$^{\clubsuit}$\thanks{\quad Equal contribution}\quad
  Masahiro Suzuki$^{\clubsuit}$\footnotemark[1]\quad
  Ryotaro Kobayashi$^{\clubsuit}$\footnotemark[1]\quad \\
  \textbf{Hiroki Sakaji}$^{\Diamond}$\quad
  \textbf{Kiyoshi Izumi}$^{\clubsuit}$ \\
  $^{\clubsuit}$The University of Tokyo, School of Engineering \\
  $^{\Diamond}$Hokkaido University, Faculty of Information Science and Technology \\
  \texttt{takayanagi-takehiro590@g.ecc.u-tokyo.ac.jp} 
}

\begin{document}
\maketitle
\begin{abstract}
Causality is fundamental in human cognition and has drawn attention in diverse research fields. 
With growing volumes of textual data, discerning causalities within text data is crucial, and causal text mining plays a pivotal role in extracting meaningful patterns.
This study conducts comprehensive evaluations of ChatGPT's causal text mining capabilities.
Firstly, we introduce a benchmark that extends beyond general English datasets, including domain-specific and non-English datasets.
We also provide an evaluation framework to ensure fair comparisons between ChatGPT and previous approaches.
Finally, our analysis outlines the limitations and future challenges in employing ChatGPT for causal text mining.
Specifically, our analysis reveals that ChatGPT serves as a good starting point for various datasets. However, when equipped with a sufficient amount of training data, previous models still surpass ChatGPT's performance.
Additionally, ChatGPT suffers from the tendency to falsely recognize non-causal sequences as causal sequences. These issues become even more pronounced with advanced versions of the model, such as GPT-4.
In addition, we highlight the constraints of ChatGPT in handling complex causality types, including both intra/inter-sentential and implicit causality. The model also faces challenges with effectively leveraging in-context learning and domain adaptation.
We release our code to support further research and development in this field.
\end{abstract}

\section{Introduction}

As causality lies at the heart of human cognition and reasoning, the study of causality is pervasive in diverse research fields including natural language processing (NLP)~\cite{girju2002causal,Pearl-09,feder-etal-2022-causal}.
When it comes to causality in text data, it provides context and meaning, connecting separate pieces of information and creating narratives.
In the digital era, where vast amounts of textual data are generated daily, understanding causalities and extracting meaningful patterns from such extensive data becomes increasingly important.

Causal text mining (CTM) involves identifying and extracting causal relations within text data. Given its crucial role in discerning patterns from the expanding pool of text data, research in this area is expanding with regard to NLP applications, models, and datasets.
Recently CTM has been applied to various NLP tasks such as knowledge base construction, question answering, and text summarization~\citep{Radinsky2013,Hassanzadeh2019, dunietz2020,Heindorf2020,Li2021}. 
With regard to the models, research has transitioned from knowledge-based systems to deep learning techniques, with a recent emphasis on encoder-based models like Bidirectional Encoder Representations from Transformers (BERT)~\citep{yang2022survey,devlin2019}. 
Additionally, there's a growing effort to create domain-specific corpora, particularly in domains like finance and medicine, and to extend research into non-English languages~\citep{chakravarthy-etal-2020-domino,gupta-2020-finlp,Akkasi2021,mariko-etal-2021-financial}.

Accordingly, efforts have been made to benchmark the growing number of models and datasets in CTM.
A notable benchmark in this trajectory is Unicausal~\citep{fiona2023}, which organizes six English datasets in a general domain and evaluates them with BERT models. 
However, there is a clear gap in addressing both domain-specific corpora and non-English datasets, underscoring the necessity for a broader benchmark and resources in this field.

On the other hand, large language models (LLMs), especially ChatGPT, have demonstrated impressive performance across numerous NLP tasks with zero-shot or few-shot in-context learning without requiring supervised training~\citep{anil2023palm2,openai2023gpt4,touvron2023llama2}. While some studies have explored the application of LLMs to NLP tasks, including sentiment analysis~\citep{wang2023chatgpt-sentiment,zhang2023sentiment}, and causal reasoning~\citep{gao2023chatgpt}, the existing body of work either does not tackle causal text mining or lacks comprehensiveness.
Thus, it remains unclear how to leverage LLMs for CTM and how to evaluate them against previous models in CTM.

Following this line of thought, the question arose ``Can ChatGPT be effectively utilized for causal text mining?'' 
To answer the question, this work undertakes comprehensive evaluations of ChatGPT's capabilities in CTM.
First of all, we establish a comprehensive benchmark for diverse datasets, ensuring a broad coverage that encompasses not only general domain English datasets but also financial domain datasets and Japanese datasets.  
We also provide a robust evaluation framework, comprising evaluation metrics and prompts tailored for a fair comparison between LLMs and traditional encoder-based models.
Lastly, we provide an in-depth analysis that highlights current limitations and suggests future paths for causal text mining.
We release our codes and resources publically available\footnote{ \url{https://github.com/retarfi/gemcausal}}, ensuring the research community can benefit and build upon our work.

Our findings offer a novel perspective on the potential of deploying ChatGPT for CTM. 
Specifically, ChatGPT often demonstrates competitive results in few-shot settings even in financial domain-specific datasets and Japanese datasets, even though a fully trained encoder-based model outperforms ChatGPT. The result indicates that ChatGPT is a good starting point for various datasets especially when training data are unavailable, but not a good causal text miner when the training data are readily available.  

In addition, our in-depth analysis highlights challenges in utilizing ChatGPT for CTM.
ChatGPT suffers from the tendency to falsely recognize non-causal sequences as causal sequences, and this problem amplifies with the more advanced models such as GPT-4.
Our findings also point out that ChatGPT struggles with complex causality types, especially those of intra/inter-sentential and implicit causality. 
Furthermore, the model also faces challenges with effectively leveraging in-context learning and domain adaptation. 

To summarize, in response to the question ``Can ChatGPT be effectively utilized for causal text mining?'', we conduct a comprehensive evaluation of ChatGPT's capabilities in CTM. Our contributions are threefold. 
Firstly, we establish a comprehensive benchmark and resources tailored for CTM, delving into the potential practicality of ChatGPT. Secondly, we introduce an evaluation framework to enable a fair comparison between ChatGPT and encoder models. Finally, through our in-depth analysis, we outline the limitations and future challenges of employing ChatGPT for CTM.

\section{Related Work}
Causal text mining, the process of extracting causal relations and information from text, expands research into various domains and develops tailored models~\cite{fiona2023}. 
With regard to domains, as different domains present unique challenges, researchers have tailored domain-specific corpora and models, such as in finance~\cite{izumi2019} and biomedical~\cite{Akkasi2021} where domain-specific knowledge is especially needed to understand causality.
Recently, the methodologies underpinning causal text mining have leveraged transformer-based encoder models~\cite{yang2022survey}. 

While encoder models enhance causal text mining performance, they rely heavily on domain-specific annotated data. The limited availability of diverse causal corpora and inconsistent annotation guidelines exacerbate the problem~\cite{fiona2023}.

In contrast, large language models (LLMs) such as ChatGPT do not require training data~\citep{ouyang2022instruct,wei2022flan}, providing a new paradigm for NLP tasks including causal text mining. 

{\tabcolsep=2.6pt
\begin{table*}
    \centering
    \small
    \renewcommand{\arraystretch}{1.1}
    \begin{tabularx}{\textwidth}{l c c c c c c}
        \toprule
        \noalign{\vskip -0.3ex}
        Datasets & Abbr. & Size & Pos:Neg & Linguistic & Inter-sent & Arguments \\
        \noalign{\vskip -0.3ex}
        \midrule
        AltLex{\footnotesize ~\citep{hidey-mckeown-2016-identifying}} & AltLex & 978 & 415:563 & AltLex & No &  -\\
        BECauSE V2.0{\footnotesize ~\citep{dunietz-etal-2017-corpus}} & BECauSE & 954 & 761:193 & Explicit & No & Phrases\\
        \begin{tabular}{@{}l}Causal-TimeBank\\{\footnotesize ~\citep{mirza-tonelli-2014-analysis,mirza-etal-2014-annotating}}\end{tabular} & CTB & 2201 & 276:1925 & Explicit & Yes & Words\\
        \begin{tabular}{@{}l}EventStoryLine\\{\footnotesize ~\citep{caselli-vossen-2016-storyline, caselli-vossen-2017-event}}\end{tabular} & ESL & 2232 & 1156:1076 & All & Yes & Words\\
        \begin{tabular}{@{}l}Penn Discourse Treebank V3.0\\{\footnotesize ~\citep{PDTB}}\end{tabular} & PDTB & 42850 & 11064:31786 & All & Yes & Clauses \\
        SemEval-2010 Task8 {\footnotesize ~\citep{hendrickx-etal-2010-semeval}} & SemEval & 10690 & 1327:9363 & All & No & Phrases \\
        FinCausal 2020{\footnotesize ~\citep{mariko-etal-2020-financial}} & FinCausal & 21325 & 1547:19778 & All & Yes & Clauses \\
        \begin{tabular}{@{}l}Japanese financial statement summaries\\{\footnotesize ~\citep{sakaji-etal-2017,kobayashi-etal-2023}}\end{tabular} & JFS & 1958 & 1429:529 & Explicit & Yes & Clauses \\
        Nikkei news articles{\footnotesize ~\citep{sakaji-etal-2017}} & Nikkei & 2045 & 898:1147 & Explicit & Yes & Clauses \\
        \bottomrule
    \end{tabularx}
    \caption[Dataset Details]{Dataset details in terms of abbreviation (Abbr.), size (Size), sequences with and without causality (Pos:Neg), linguistic indicators of causal relations (Linguistic), treatment of inter-sentential causality (Inter-sent), and the length of causal arguments annotated (Arguments).}
    \label{table:tab_dataset}
\end{table*}
}

Accordingly, there has been significant interest in evaluating ChatGPT's causal abilities in recent years.
\citet{gao2023chatgpt} investigated ChatGPT's performance across multiple causal reasoning tasks.  
Similarly, both \citet{qin2023chatgpt} and \citet{chan2023chatgpt} have assessed ChatGPT's causal reasoning capacities, with \citet{kiciman2023causal} specifically focusing on the model's causal discovery skills.
Moreover, \citet{tu2023causal} have evaluated ChatGPT's causal reasoning in medical domain where specialized knowledge is essential.


\newpage
\section{Evaluation Settings}

\subsection{Datasets}
\label{sec:dataset}
For comprehensive evaluations of CTM, we organize a variety of datasets related to causal text mining across different domains and languages.

The characteristics of the datasets in our experiments are summarized in Table~\ref{table:tab_dataset}. 
In addition to data size and the ratio of positive and negative samples, we organize the datasets based on the different causality types. 
Explicit causality involves direct statements using causal connectives or phrases like ``because'' or ``as a result,'' while implicit causality refers to the representation of causal relations without obvious markers~\citep{dunietz-etal-2017-corpus}.
 In the Linguistic column, AltLex denotes alternative lexicalization connectives, which are linguistically broader than explicit ones~\citep{hidey-mckeown-2016-identifying}, and ``All'' encompasses all linguistic patterns, including implicit causality.
In addition, the causes and the effects are contained within a sentence in intra-sentential causality, while inter-sentential causality stretches over several sentences.
In Table~\ref{table:causality-type-examples}, we present examples of the texts addressed in this study and their corresponding causality types.

\begin{table*}
    \centering
    \renewcommand{\arraystretch}{1.4}
    \begin{tabularx}{\textwidth}{Xl}
        \toprule
        \noalign{\vskip -0.7ex}
        Annotated Sequences & Causality Types  \\
        \noalign{\vskip -0.7ex}
        \midrule
        \footnotesize <e>Investors may have reacted so strongly to Friday's U.S. stock market loss</e> \textit{because} <c>they had vivid memories of the Frankfurt exchange's losing 35\% of its value in the 1987 crash and its wake</c>. & \small Explicit \& Intra-sentential \\
        \footnotesize <e>South Korea registered a trade deficit of \$101 million in October</e>, <c>reflecting the country's economic sluggishness</c>, according to government figures released Wednesday. & \small Implicit \& Intra-sentential \\
        \footnotesize <c>Propaganda is just information to support a viewpoint</c>, and the beauty of a democracy is that it enables you to hear or read every viewpoint and then make up your own mind on an issue. <e>Government press releases, speeches, briefings, tours of military facilities, publications are all propaganda of sorts</e>. & \small Implicit \& Inter-sentential \\
        \bottomrule
        \end{tabularx}
    \caption{Examples of sequences and their causality types. The italicized part is the marker for explicit causal relations. The <c> and <e> tags indicate cause and effect spans. All examples are sourced from the Penn Discourse Treebank V3.0~\protect\citep{PDTB}.}
    \label{table:causality-type-examples}
\end{table*}

\begin{figure*}[ht]
  \begin{center}
    \includegraphics[clip,keepaspectratio,width=\linewidth]{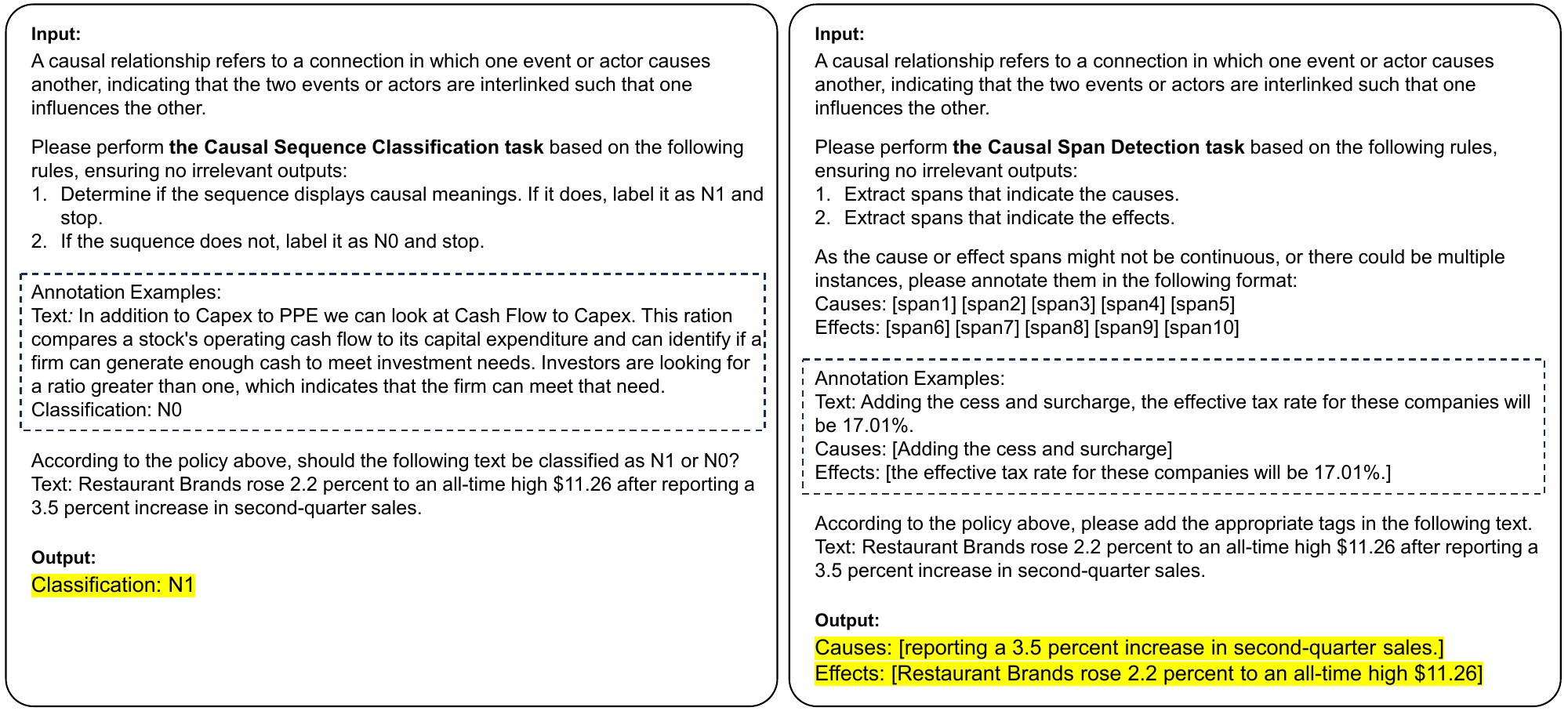}
    \caption{Illustrations of prompts for the causal sequence classification task and the causal span detection task. Regions outlined by dotted lines represent demonstrations of the few-shot setting and are omitted under the zero-shot setting. All text examples are sourced from the FinCausal 2020 dataset~\cite{mariko-etal-2020-financial}.}
    \label{fig:prompt-examples}
  \end{center}
\end{figure*}

\subsection{Investigated Tasks}\label{sec:investigated_tasks}
The methodologies of CTM often involve two phases: causal sequence classification and causal span detection~\cite{mariko-etal-2020-financial, mariko-etal-2021-financial, mariko-etal-2022-financial, Tan-22-causal}.
The causal sequence classification is a binary classification task to detect whether the sequence entails causality or not.
This task requires a deep understanding of commonsense knowledge, as determining causality necessitates the comprehension of underlying real-world principles and contexts~\cite{gao2023chatgpt}.

The causal span detection task aims to distinguish between cause and effect arguments present in causal sequences.
This task requires a precise understanding of a complex context that comprises multiple entities and events to discern which parts of sequences correspond to causes and effects and which are noise, in addition to the capabilities previously mentioned.

We experiment on all datasets in Table~\ref{table:tab_dataset} for causal sequence classification. For causal span detection, we experiment on AltLex, BECauSE, PDTB, FinCausal, and JFS 
because the annotated causal arguments are sufficiently long and appropriate for the task.

\subsection{Evaluation Framework}
We introduce an evaluation framework designed to enable fair comparisons between LLMs and prior encoder-based models in the context of CTM.

For causal sequence classification, we adopt a binary classification setup illustrated in Figure~\ref{fig:prompt-examples}, because ChatGPT as well as encoder models has consistently demonstrated proficiency in binary classifications, similar to the observations in past research~\cite{zhang2023sentiment}. We evaluate the performance using metrics such as binary F1, accuracy, precision, and recall, following previous studies in CTM~\cite{yang2022survey,fiona2023}.

With regard to causal span detection, we introduce SQuAD-styled evaluation methods, inspired by the Stanford Question Answering Dataset (SQuAD)~\citep{rajpurkar-etal-2016-squad}. 
This approach stands in contrast to traditional CTM research~\citep{mariko-etal-2020-financial,mariko-etal-2021-financial,mariko-etal-2022-financial,fiona2023}, which predominantly relies on a named entity recognition (NER) style framework. 
In the latter traditional approach, models label each token with a tag (e.g., ``C'' for Cause, ``E'' for Effect, and ``-'' for others). 
However, leveraging ChatGPT for tokenizing sentences and outputting token-specific tags yielded suboptimal results due to inaccurate segmentations, bringing about “word misalignments”.
For instance, the phrase ``It'll cost \$15'' would be segmented into ``It, 'll, cost, \$, 15''.\ Such tokenization challenges distort the fair evaluations of ChatGPT from causal text mining, as they grapple with understanding proper alignment and word segmentation.

In light of these observations, our study emphasizes the extraction of both causes and effects in the causal span detection task as illustrated in Figure~\ref{fig:prompt-examples}.
In our evaluation method, the model extracts specific word spans and assesses its performance using metrics such as exact match, macro F1. 
Specifically, the exact match measures the percentage of predictions that match the ground truth answers of both cause and effect exactly. 
Macro F1, precision, and recall are calculated by assessing overlaps between predictions and ground truth for each cause and effect.
These metrics are calculated on a per-word basis for models like ChatGPT, and on a per-subword token basis for encoder models.
We then average metrics for both cause and effect across all samples.

To maintain a consistent and balanced evaluation across the different tasks and datasets, we have limited our sampling to a maximum of 500 examples from the test section of each dataset.

\begin{table*}[h]
    \centering
    \begin{tabular}{lcccccccccc}
    \toprule
     & \multicolumn{3}{c}{ChatGPT (0shot)} & \multicolumn{3}{c}{GPT-4 (0shot)} & \multicolumn{3}{c}{DeBERTaV3} \\
    \cmidrule(lr){2-4} \cmidrule(lr){5-7} \cmidrule(lr){8-10}
    Dataset & P & R & F1 & P & R & F1 & P & R & F1 \\
    \midrule
    AltLex & .371 & .809 & .508 & .381 & .887 & .533 & .537 & .765 & \textbf{.631} \\
    BECauSE & .957 & .537 & .688 & .848 & .683 & .757 & .820 & 1.000 & \textbf{.901} \\
    CTB & .162 & .405 & .231 & .169 & .905 & .285 & .714 & .476 & \textbf{.571} \\
    ESL & .504 & .593 & .545 & .510 & .894 & .650 & .857 & .690 & \textbf{.765} \\
    PDTB & .128 & .487 & .203 & .336 & .841 & .480 & .770 & .740 & \textbf{.755} \\
    SemEval & .848 & .262 & .400 & .234 & .863 & .368 & .875 & .961 & \textbf{.916} \\
    FinCausal & .488 & .500 & .494 & .120 & .769 & .208 & .646 & .795 & \textbf{.713} \\
    Nikkei & .532 & .515 & .523 & .503 & .925 & .652 & .908 & .988 & \textbf{.946} \\
    JFS & .280 & .784 & .412 & .782 & .913 & .842 & .906 & .966 & \textbf{.935} \\
    \midrule
    \midrule
    & FA & PA & NA & FA & PA & NA & FA & PA & NA \\
    \cmidrule(lr){2-4} \cmidrule(lr){5-7} \cmidrule(lr){8-10}
    AltLex & .551 & .809 & .448 & .554 & .887 & .420 & .743 & .765 & .734 \\
    BECauSE & .608 & .537 & .900 & .647 & .683 & .500 & .824 & 1.000 & .100 \\
    CTB & .642 & .405 & .679 & .396 & .905 & .318 & .905 & .476 & .971 \\
    ESL & .517 & .593 & .445 & .530 & .894 & .185 & .793 & .690 & .891 \\
    PDTB & .702 & .487 & .720 & .472 & .841 & .321 & .900 & .740 & .942 \\
    SemEval & .403 & .262 & .851 & .698 & .863 & .679 & .982 & .961 & .984 \\
    FinCausal & .600 & .500 & .664 & .542 & .769 & .523 & .950 & .795 & .963 \\
    Nikkei & .574 & .515 & .623 & .615 & .925 & .416 & .956 & .988 & .936 \\
    JFS & .772 & .784 & .771 & .740 & .913 & .191 & .898 & .966 & .681 \\
    \bottomrule
    \end{tabular}
    \caption{Experimental results on the causal sequence classification task. P, R and F1 indicate Precision, Recall and binary F1-score, respectively. 
    FA, PA, and NA indicate accuracy on the full test data, causal sequence, and non-causal sequence, respectively.}
    \label{tab:result_sequence_classification}
\end{table*}

\begin{table*}
    \centering
    \begin{tabular}{lcccccccccccccccc}
    \toprule
     & \multicolumn{4}{c}{ChatGPT (0shot)} & \multicolumn{4}{c}{GPT-4 (0shot)} & \multicolumn{4}{c}{DeBERTaV3} \\
    \cmidrule(lr){2-5} \cmidrule(lr){6-9} \cmidrule(lr){10-13}
    Dataset & EM & F1 & P & R & EM & F1 & P & R & EM & F1 & P & R \\
    \midrule
    AltLex & .000 & .488 & .605 & .457 & .000 & .527 & .589 & .546 & \textbf{.162} & \textbf{.700} & .700 & .704 \\
    BECauSE & .025 & .466 & .494 & .515 & .025 & .468 & .434 & .637 & \textbf{.350} & \textbf{.831} & .813 & .854 \\
    PDTB & .034 & .471 & .519 & .493 & .032 & .502 & .495 & .615 & \textbf{.482} & \textbf{.856} & .867 & .845 \\
    FinCausal & .015 & .531 & .637 & .508 & .029 & .666 & .711 & .694 & \textbf{.706} & \textbf{.901} & .895 & .908 \\
    JFS & .000 & .704 & .759 & .710 & .014 & .778 & .799 & .831 & \textbf{.493} & \textbf{.935} & .938 & .932 \\
    \bottomrule
    \end{tabular}
    \caption{Experimental results on the causal span detection task. P, R and F1 indicate Precision, Recall and macro F1-score, respectively. EM stands for an exact match.}
    \label{tab:result_span_detection}    
\end{table*}

\subsection{Investigated Models}
We utilize the latest versions of \texttt{gpt-3.5-turbo} and \texttt{gpt-4} for the ChatGPT family models.
In the case where the input reaches the maximum length limit in many shots setting, we switch to \texttt{gpt-3.5-turbo-16k} due to the approximately 4k maximum length constraint of \texttt{gpt-3.5-turbo}.

Recent works in CTM ~\citep{mariko-etal-2020-financial,mariko-etal-2021-financial,mariko-etal-2022-financial,fiona2023} increasingly utilize Transformer encoders such as BERT~\citep{devlin2019} or RoBERTa~\citep{liu2019roberta}.
Accordingly, we employ DeBERTaV3~\citep{he2023debertav3} as the best baseline model of encoders for the English datasets.
For a concise comparison, we also use a plain configuration of DeBERTaV3 with only a multilayer perceptron (MLP) connected for classification.
For the Japanese datasets, we employ Japanese BERT\footnote{\url{https://huggingface.co/cl-tohoku/bert-base-japanese}} as a baseline model.

\subsection{Prompting Strategy}
Prior investigations indicate that the performance of LLMs is sensitive to prompts and may result in very different responses even when using semantically similar prompts~\cite{Ethan2022,lu-etal-2022-fantastically}. 

Considering the model's sensitivity to the prompts, we provide coherent prompts across tasks and datasets to ensure consistent evaluations for CTM in our experiments. 
Following the previous work~\cite{zhang2023sentiment}, our prompt consists of a causality description, task definition, task name, and output format as illustrated in Figure~\ref{fig:prompt-examples}.
An additional demonstration part contains $k$ instances with their corresponding labels formatted appropriately. 
Throughout our experiments, $k$ instances are randomly selected for demonstrations in in-context learning and shared across all test cases.

\section{Experimental Results}
\label{sec:result}
Table~\ref{tab:result_sequence_classification} presents the results of the causal sequence classification.
DeBERTaV3 consistently surpassed both ChatGPT and GPT-4 across most metrics, while ChatGPT performs relatively well regardless of domains and languages in zero-shot settings.
The result indicates that ChatGPT serves as a good starting point when training data are limited as its performance is not influenced by the data size.
In contrast, encoder models depend heavily on data size~\footnote{In our results, we calculated correlation coefficients between F1 scores and the log of training data size. We found that ChatGPT has lower correlation coefficients (0.02 and -0.15), while DeBERTaV3 has higher coefficients (0.58 and 0.44) in sequence classification and span detection tasks.}~\cite{mehrafarin-etal-2022-importance} and struggle to generalize across CTM datasets because of varied causal definitions and linguistic coverage~\cite{fiona2023}.



In terms of the comparison between GPT variants, GPT-4 demonstrated a consistent decline in negative accuracy across all datasets compared to ChatGPT. 
The result stems from GPT-4's tendency to incorrectly categorize non-causal sequences as causal. 
This ``causal hallucination'' becomes also evident in its lower precision on datasets such as CTB, SemEval, and FinCausal, which predominantly consist of negative samples as shown in Table~\ref{table:tab_dataset}. 

Lastly, the Japanese dataset, especially JFS, demonstrated high performance across models due to the formality of its source, financial statements. These documents, like earnings summaries, present logic clearly, making causality identification easier. In contrast, ChatGPT's weaker performance highlights difficulties in understanding financial text nuances in zero-shot scenarios.



We summarize the results of causal span detection in Table~\ref{tab:result_span_detection}.
DeBERTaV3 also outperformed both ChatGPT and GPT-4. ChatGPT, in particular, struggled remarkably in the exact match metric. We believe this discrepancy in exact match performance arises from varying annotation styles of arguments across datasets, as mentioned in Table~\ref{table:tab_dataset}.
The variability of annotations requires adaptive capabilities from the model, which becomes particularly challenging in zero-shot scenarios.

A comparative analysis between ChatGPT and GPT-4 reveals the latter's higher performance in most datasets across evaluation metrics.  The causal hallucination issues observed in GPT-4 during causal sequence classification seem to be less severe in the cause-effect span detection task. This could be attributed to the task's objective of extracting causal segments, which leaves limited room for any hallucinations.

Lastly, in the context of the performance of the Japanese dataset, similar to the results in causal sequence classification, the Japanese dataset, particularly JFS, demonstrated high performance across all models. In addition to the formality of the dataset, we point out that punctuations are located near clue expressions, such as ``lead to'' due to the formality of the text.
This characteristic could help models capture the clear spans from the sequences.

\begin{figure*}[t]
    \centering
    \includegraphics[width=\linewidth]{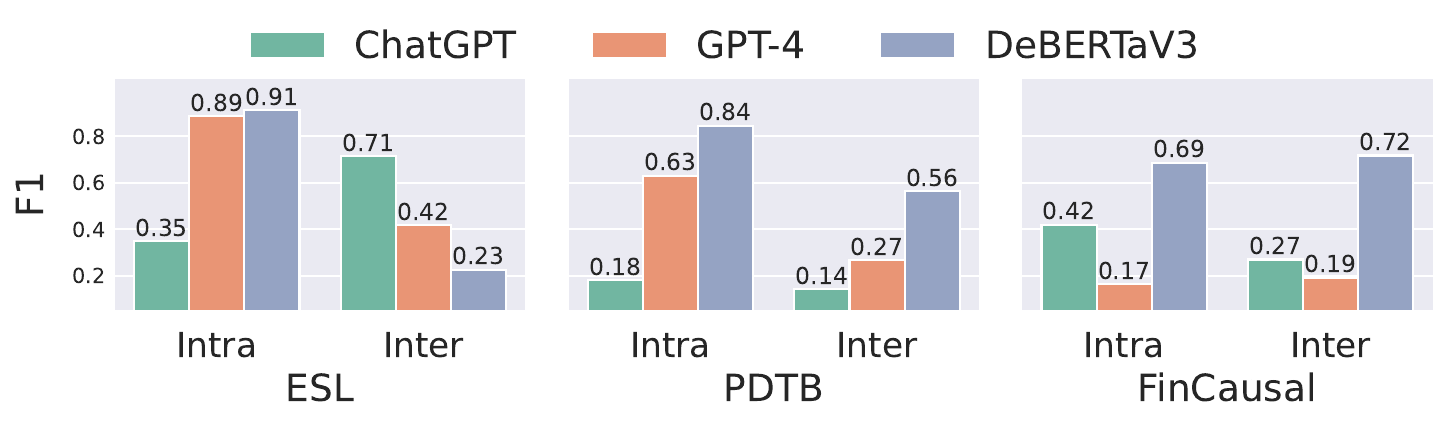}
    \caption{Performance comparisons for inter and intra-sentential causality in the causal sequence classification task on the ESL, PDTB, and Fincausal datasets, respectively.}
    \label{fig:sc_diffs}
\end{figure*}

\begin{figure*}[h]
    \centering
    \includegraphics[width=\linewidth]{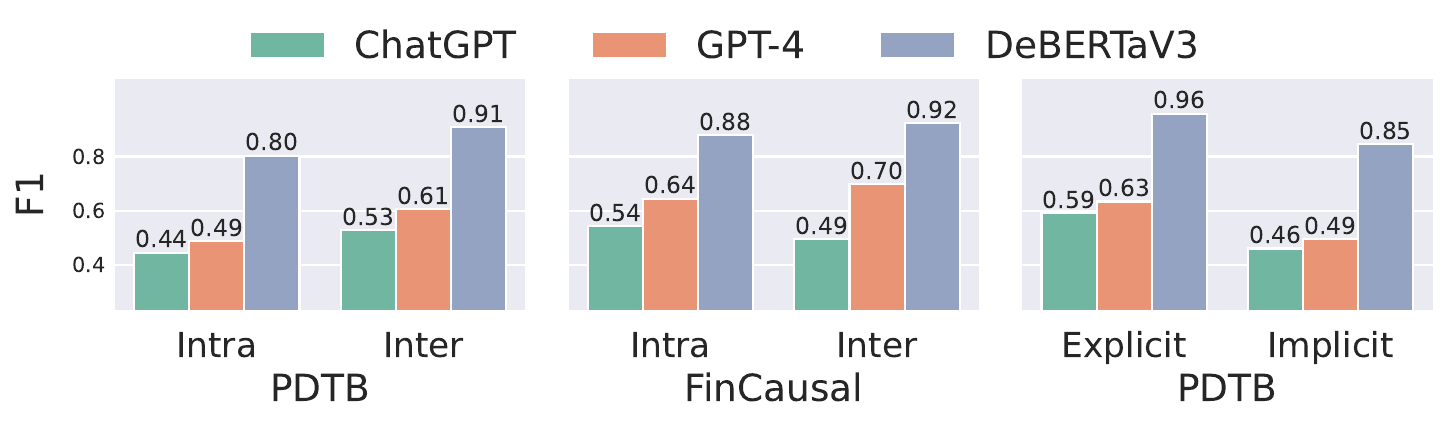}
    \caption{Performance comparisons for different causality types in causal span detection tasks. The left two bar plots represent performance for inter and intra-sentential causality on the PDTB and FinCausal datasets, while the right plot illustrates performance for explicit and implicit causality on the PDTB dataset. }
    \label{fig:sd_diffs}    
\end{figure*}

\section{Analysis}
We proceed to conduct an in-depth analysis for the comprehensive evaluations of ChatGPT in various scenarios. 
Specifically, we begin by examining ChatGPT's performance for the different types of causality: intra/inter-sentential and explicit/implicit. Subsequently, we study the influence of in-context learning in causal text mining, adjusting the number of demonstrations provided. We then focus our analysis on domain-specific corpora, especially the financial domain, where specific domain knowledge is important to understand causality.

\subsection{Different types of causality}
As previously detailed in Section~\ref{sec:investigated_tasks}, causal text mining tasks encompass various causality types.
We investigate the impact of different causality types on causal text mining tasks, specifically inter/intra sentential, and explicit/implicit causality.

As described in Section~\ref{sec:dataset}, intra-sentential causality encompasses cause and effect expressions in one sentence, the cause and effect expressions lie in different sentences.
Performance comparisons for inter and intra-sentential causality in the causal sequence classification and the causal span detection tasks are illustrated in Figure~\ref{fig:sc_diffs} and Figure~\ref{fig:sd_diffs}, respectively.

Figure~\ref{fig:sc_diffs} shows that most datasets exhibit superior intra-sentential causal sequence classification performance. 
The challenge of inter-sentential causality, characterized by extended sequence and greater entanglement of causal events and noises, explains the result.

Conversely, as illustrated in Figure~\ref{fig:sd_diffs}, most datasets demonstrate higher performance in inter-sentential causal span detection. This is likely because, in inter-sentential cases, the cause and effect spans are often neatly located in separate sentences, often bouned by either periods or punctuations. This makes it easier for models to identify and extract them. 

The distinction between explicit and implicit causality is another important aspect of different types of causality. 
To distinguish between explicit and implicit causality in the PDTB datasets\footnote{We excluded the Fincausal dataset due to its limited explicit causality samples.}, we employed explicit expressions listed in the GitHub repository for the BECauSE dataset\footnote{\url{https://github.com/duncanka/BECauSE}}.

Figure~\ref{fig:sd_diffs} depicts the performance comparison between explicit and implicit causality for span detection in the PDTB dataset. The results show both ChatGPT and DeBERTaV3 face more difficulties in extracting spans from implicitly causal sentences.
The result suggests that the nuances present in implicit causality demand a deeper comprehension and more sophisticated approaches to comprehending causality, compared to the straightforward nature of explicit causality where causal markers are overtly present.

\begin{figure}[tb] 
    \centering
    \includegraphics[width=\linewidth]{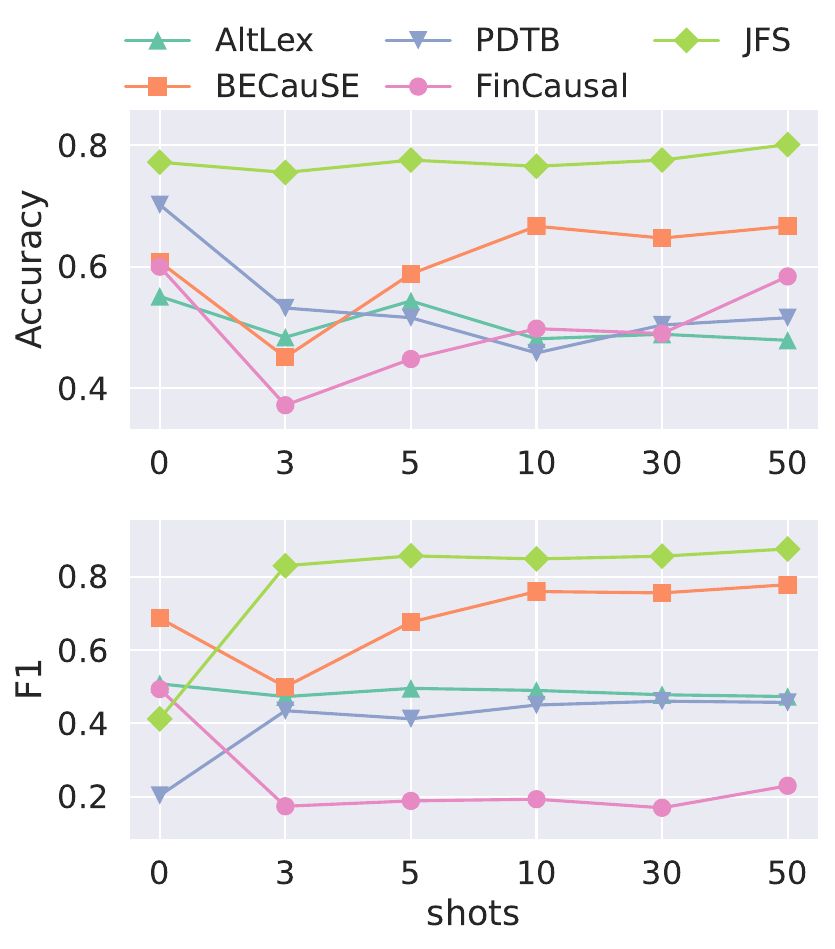}
    \caption{Performance of ChatGPT on causal sequence classification across multiple datasets with varying number of demonstrations.}
    \label{fig:shots_sc}
\end{figure}

\begin{figure}[tb] 
    \centering
    \includegraphics[width=\linewidth]{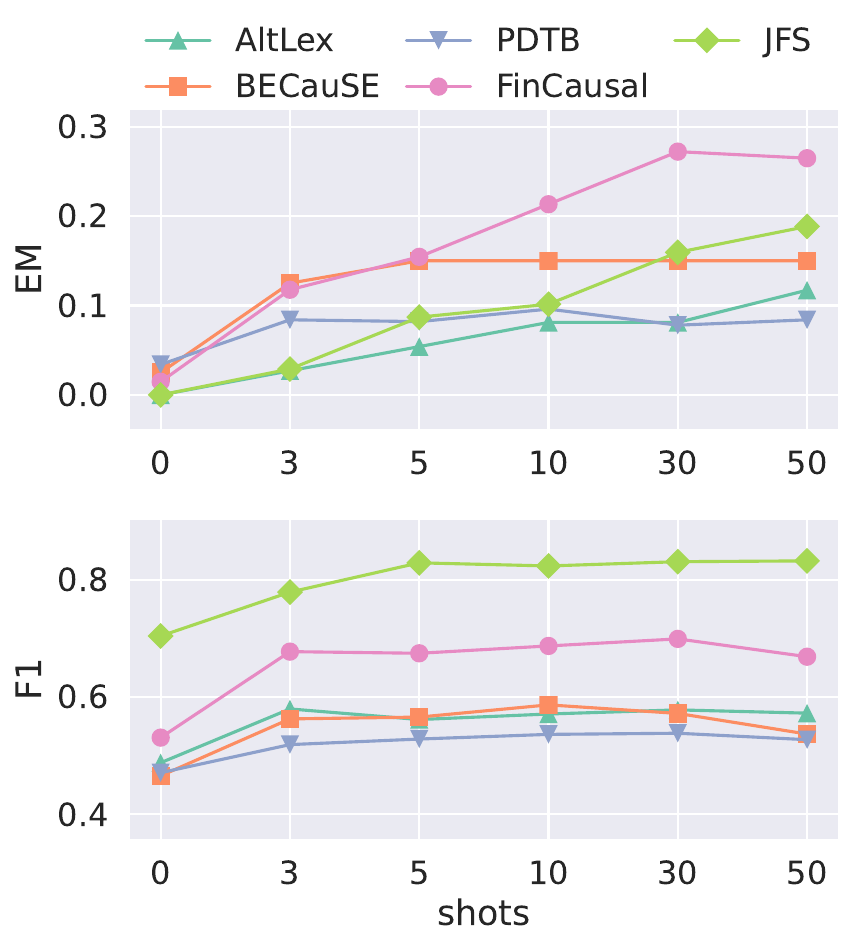}
    \caption{Performance of ChatGPT on causal span detection across multiple datasets with varying number of demonstrations.}
    \label{fig:shots_sd}
\end{figure}

\subsection{In context learning}
We further examine how ChatGPT's performance is influenced by varying the number of demonstrations provided. 
Results for the causal sequence classification are depicted in Figure~\ref{fig:shots_sc}, while the results for cause-effect span detection is illustrated in Figure~\ref{fig:shots_sd}.

From Figure~\ref{fig:shots_sc}, it is evident that providing ChatGPT with additional shots only yields marginal improvements in performance. 
We consider the simplistic binary labels (0,1) in our causal sequence classification tasks might not sufficiently inform ChatGPT about the underlying causal reasons by just adding demonstrations in the current form. As a result, there's a need for future research to explore more effective ways of guiding ChatGPT through demonstrations.

In contrast, Figure~\ref{fig:shots_sd} shows that with the increasing number of demonstrations, there's a pronounced improvement in the span detection tasks. This result can be understood through two main considerations. 
Firstly, the wide range of annotation styles presented in Table~\ref{table:tab_dataset} underscores the benefit of multiple demonstrations as understanding the nuances of the annotation guidelines directly determines the performance of cause-effect span detection. By introducing ChatGPT to these varied annotations examples, the model gradually refines its understanding towards the characteristics of the dataset.

Secondly, previous studies such as~\citet{zhang2023sentiment} indicate that in-context learning is more effective for a difficult task. 
The causal detection is inherently more challenging than mere sequence classification. Instead of straightforward binary outcomes, the model is tasked with precisely identifying and extracting particular text spans.
This increased complexity likely amplifies the advantages of increased demonstrations compared to the causal sequence classification.



\subsection{Doamin adaptation}


{\tabcolsep=3pt
\begin{table}
\centering
\begin{tabular}{lcccccccccccc}
\toprule
 & \multicolumn{2}{c}{Sequence} & \multicolumn{2}{c}{Span}  \\
 \cmidrule(lr){2-3} \cmidrule(lr){4-5} 
 Method & Acc & F1 & EM & F1 \\
\midrule
ChatGPT (0 shot) & .494 & .600 & .015 & .531\\
ChatGPT (3 shots) & .174 & .372 & .118 & .678\\
ChatGPT (0 shot \& AN) & .210 & .744  & .022 & .524 \\
ChatGPT (3 shots \& AN) & .176 & .344  & .125 & .668 \\
ChatGPT (0 shot \& EXP) & .204 & .734 &.007 & .514  \\
ChatGPT (3 shots \& EXP) & .176 & .344 &.132 & .673 \\
GPT-4 (0 shot) &.208& .542& .029& .666\\
GPT-4 (3 shots) &  .230 & .678 &.110&.695\\
DeBERTaV3 & \textbf{.713} & .950  & \textbf{.706} & \textbf{.901}\\
FinBERT &.700 & \textbf{.952} & .426 & .861 \\
\bottomrule
\end{tabular}
\caption{Performance comparison of various models on the Fincausal dataset for both causal sequence classification and causal span detection tasks for the financial domain. AN and EXP represent acting within the prompt as analyst and expert, respectively. The accuracy and F1 scores are reported for sequence classification, and Exact Match (EM) and F1 scores for span detection.}
\label{tab:domain_adaptation}
\end{table}
}

The research in CTM shows that domain-specific language model increases performance for domain-specific corpora such as financial documents~\citep {chakravarthy-etal-2020-domino,gupta-2020-finlp}.
We investigate the approaches for ChatGPT to adapt to domain-specific corpora using the Fincausal dataset.

For ChatGPT, we add the sentence ``you are a financial analyst'' or ``you are a financial expert'' to the prompts in order to define context and role for financial task~\cite{santu2023teler, white2023prompt}.
As for encoder models, we utilize FinBERT~\cite{araci2019finbert} as domain-specific models.
These models are further pre-trained using general BERT models with financial corpora.

The experimental results for domain adaptation are summarized in Table~\ref{tab:domain_adaptation}. The result shows that the DeBERTaV3 model surpasses FinBERT across most evaluation metrics. This superior performance can be linked to the capabilities of the base model. While DeBERTaV3 consistently outshines BERT in standard language tasks, FinBERT achieves a slight edge over the general BERT model when tailored using a financial corpus~\citep{araci2019finbert,SUZUKI2023}.

Regarding ChatGPT's performance, introducing context and roles with prompts like ``you are a financial analys'' does not lead to enhanced results. Enhancing ChatGPT's proficiency with domain-specific datasets remains a challenge and needs further exploration for domain adaptation.

\section{Conclusion}
In this work, we conduct comprehensive evaluations of ChatGPT's potential in CTM to answer the question ``Can ChatGPT be effectively utilized for causal text mining?'' 
To this end, we organize datasets spanning various domains and languages, establishing a holistic benchmark for causal text mining. 
We also provide an evaluation framework that ensures fair comparison when contrasting ChatGPT with traditional encoder-based models.
Our comprehensive evaluations reveal that ChatGPT serves as a good starting point for various datasets, including domain-specific and non-English datasets. 

Then, our in-depth analysis reveals the limitations and future work of employing ChatGPT for causal text mining. 
Our analysis highlights the constraints of ChatGPT in handling complex causality types, including both intra/inter-sentential and implicit causality. 
This demands more sophisticated approaches to comprehending causality, compared to straightforward causality. 
The model also faces challenges with effectively leveraging in-context learning and domain adaptation. As a result, there’s a need for future research to explore more effective ways of guiding ChatGPT through demonstrations and domain adaption.


\bibliography{causal_nlp}

\appendix
\section{Datasets Details}
\label{sec:datasets_details}

This section details the causal text mining dataset utilized in our study.
AltLex~\citep{hidey-mckeown-2016-identifying} investigates causal relations in English Wikipedia articles, focusing on alternative lexicalization connectives (AltLex) within a single sentence. 
These AltLex connectives appear in a broader range of linguistic forms compared to explicit markers, exemplified by phrases such as ``This may help explain why'' and ``This activity produced.'' 
However, one limitation of this dataset is its assumption that all words preceding and succeeding the causal cues are the sole spans corresponding to the cause and effect intended for extraction.

BECauSE V2.0 (BECauSE)~\citep{dunietz-etal-2017-corpus} is designed to annotate intra-sentential explicit causal relations, aiming to encapsulate the diverse constructions utilized to convey cause and effect. 
The arguments directed to causal relation instances are phrase units.
The dataset is derived from several sources, namely the New York Times Annotated Corpus (NYT)~\citep{NYT}, Penn Treebank (PTB)~\citep{Marcus-94}, Congressional Hearings of the 2014 NLP Unshared Task in PoliInformatics (CHRG)~\citep{Smith-14}, and the Manually Annotated Sub-Corpus (MASC)~\citep{MASC}.

Causal-TimeBank (CTB)~\citep{mirza-tonelli-2014-analysis,mirza-etal-2014-annotating} and EventStoryLine (ESL)~\citep{caselli-vossen-2016-storyline, caselli-vossen-2017-event} datasets are widely recognized in the field of Event Causality Identification (ECI), which concentrates on discerning causal links between events within textual data.
For instance, the ECI model identifies a causal link between ``earthquake'' and  ``tsunami'' in the sentence ``The earthquake generated a tsunami.''
The CTB dataset originates from the TimeBank corpus of the TempEval-3 task~\citep{uzzaman-etal-2013-tempeval3} and is designed to annotate solely explicit causal relations via a rule-driven algorithm. 
In contrast, ESL derives from an extended version of the EventCorefBank (ECB+)~\citep{cybulska-vossen-2014-using} and encompasses explicit and implicit causality.
Both datasets address intra-sentential as well as inter-sentential causality.
However, a limitation is evident in that only the initial words of an event are tagged, leading to the omission of the context from the extracted arguments.

Penn Discourse Treebank V3.0 (PDTB)~\citep{PDTB} represents the third installment of the Penn Discourse Treebank project and stands as the most extensive annotated corpus dedicated to discourse relations.
This project primarily focuses on annotating discourse relations present in the Wall Street Journal (WSJ) segment of Treebank-2.
The uniqueness of PDTB lies in its ability to annotate not only overt discourse relations with explicit connectives but also those that are conveyed through varied forms containing inter-sentential and implicit causality.
It is pertinent to note that causal relations within clauses are excluded from annotation.
Although causal relation is not the main focus of the dataset, one of the relations annotated by PDTB is treated as causality in our study.

In the SemEval-2010 Task8 (SemEval)~\citep{hendrickx-etal-2010-semeval}, the primary emphasis is not causal relations but the multi-faceted classification of semantic relations between noun phrase pairs.
The dataset restricts relation instances to those present within a single sentence; however, they are not restricted to merely explicit instances.
In the nine relations defined by SemEval, we treat the ``Cause-Effect (CE)'' relation as a causal relation.
We select the six datasets mentioned above based on the previous study related to causal text mining~\citep{fiona2023}.

FinCausal 2020 (FinCausal)~\citep{mariko-etal-2020-financial} dataset, extracted from financial news articles published in 2019 by Quam and additional data from the EDGAR Database of the U.S. Securities and Exchange Commission (SEC), belongs to the financial domain.
This dataset is made for the FinCausal-2020 Shared Task on ``Financial Document Causality Detection,'' which aims to develop the ability to use external information to explain why changes in market and corporate financial conditions occur.
Key features of the FinCausal 2020 dataset are its restriction of effect spans to quantifiable facts and the common occurrence of causal arguments being presented as entire sentences.
The dataset is also annotated for implicit and inter-sentential causality.

Our study also employs Japanese datasets sourced from Japanese financial statement summaries (JFS) and Nikkei news articles (Nikkei)~\citep{sakaji-etal-2017,kobayashi-etal-2023}.
JFS are mandated disclosure documents for publicly listed companies, providing details on business performance and financial condition, typically characterized by specialized and standardized phrasing.
Nikkei refers to a financial newspaper published by Nikkei, Inc.
These datasets analyze causal relations present within a single sentence or spanning two adjacent sentences using explicit markers derived from hand-labeling tags on 300 newspaper articles focused on economic trends.
In addition to datasets labeled for the presence or absence of causality in sentences, another dataset from JFS is annotated for cause and effect spans by an investor with 15 years of experience.
From examining 30 files in the latter dataset, 478 causal relations are identified.

\end{document}